# Space Physiology and Technology: Physical Adaptations, Countermeasures, and Opportunities for Wearable Systems

Shamas U.E. Khan[1†], Rejin J. Varghese[1†], Panagiotis Kassanos[1], Dario Farina[1*], and Etienne Burdet[1*]

[1]Department of Bioengineering, Imperial College London, London, United Kingdom.
[*]Address correspondence to: {d.farina, e.burdet}@imperial.ac.uk
[†]These authors contributed equally to this work.

## Abstract

Space poses significant challenges for humans, leading to physiological adaptations in response to an environment vastly different from Earth. A comprehensive understanding of these physiological adaptations is needed to devise effective countermeasures to support human life in space. This narrative review first focuses on the impact of the environment in space on the musculoskeletal system. It highlights the complex interplay between bone and muscle adaptations and their implications on astronaut health.

Despite advances in current countermeasures, such as resistive exercise and pharmacological interventions, they remain partially effective, bulky, and resource-intensive, posing challenges for future missions aboard compact spacecraft. This review proposes wearable sensing and robotic technology as a promising alternative to overcome these limitations. Wearable systems, such as sensor-integrated suits and (soft) exoskeletons, can provide real-time monitoring, dynamic loading, and exercise protocols tailored to individual needs. These systems are lightweight, modular, and capable of operating in confined environments, making them ideal for long-duration missions.

In addition to space applications, wearable technologies hold significant promise for terrestrial uses, supporting rehabilitation and assistance for the ageing population, individuals with musculoskeletal disorders, and enhance physical performance in healthy users. By integrating advanced materials, sensors and actuators, and intelligent and energy-efficient control, these technologies can bridge gaps in current countermeasures while offering broader applications on Earth.

## 1. Introduction

Human physiology continually adapts to the Earth's gravitational field from embryonic development. As a result, gravitational unloading in outer space causes adverse effects that challenge human space exploration and habitation [1]. Frequent space missions and experiments at the International Space Station (ISS) (orbiting 300-435 km above Earth) have provided extensive data on the physiological challenges faced by astronauts [2] (Fig.1). As astronauts ascend from Earth, a headward bodily fluid shift occurs [3], causing changes in body fluid distribution and electrolyte homeostasis. Muscle atrophy and bone resorption are significant concerns during prolonged stays in space. Early space missions lasting just a few days resulted in muscle atrophy of up to 16%, despite countermeasures [4]. Bone resorption occurred at a rate of 1-2% per month, and though researchers partially reduced the effects with countermeasures, they were not fully mitigated.

The human musculoskeletal system allows for locomotion, postural control, and performing activities of daily living. In space missions, adverse effects on this system affect task performance, increase long-term risks, and require intense post-flight rehabilitation [4]. Astronauts may often travel to space for several months, and the adaptations observed in human physiology post-flight resemble that of ageing on Earth [5]. With longer space missions (and prolonged exposure to radiation and microgravity), the risk becomes more significant and could even prove fatal if not mitigated. It is uncertain how long life can be sustained in microgravity, but effective countermeasures for space-related challenges are imperative for safe, extended space travel.

Through this review, we discuss the challenges of the outer space environment and the musculoskeletal health complications derived from sustained exposure. The currently employed solutions to counteract these complications are discussed, as well as their limitations, and the potential solutions to these challenges. The paper is organized as follows. First, outer space's risks and stressors that impact the musculoskeletal system are discussed (Section 2), as well as the impact that these have to the musculoskeletal system (Section 3). This is followed by a discussion on the currently deployed and research-based countermeasures adopted by different space agencies and laboratories, and the limitations of these solutions (Section 4). The opportunity for wearable robotic and sensing technologies is then highlighted as a potential alternative to current exercise-based countermeasures in Section 5 along with recent examples from the literature. Finally, a discussion section (Section 6) concludes the paper by analysing the main takeaway messages of the paper, various shortcomings of studies conducted thus far, and highlights focus areas for further research and innovation.

## 2. Outer Space Stressors

Spaceflight exposes astronauts to multiple stressors, leading to adverse physiological adaptations in almost all organ systems [6]. The most prominent stressors to the musculoskeletal system - weightlessness/microgravity, radiation, and psychosocial factors are discussed here.

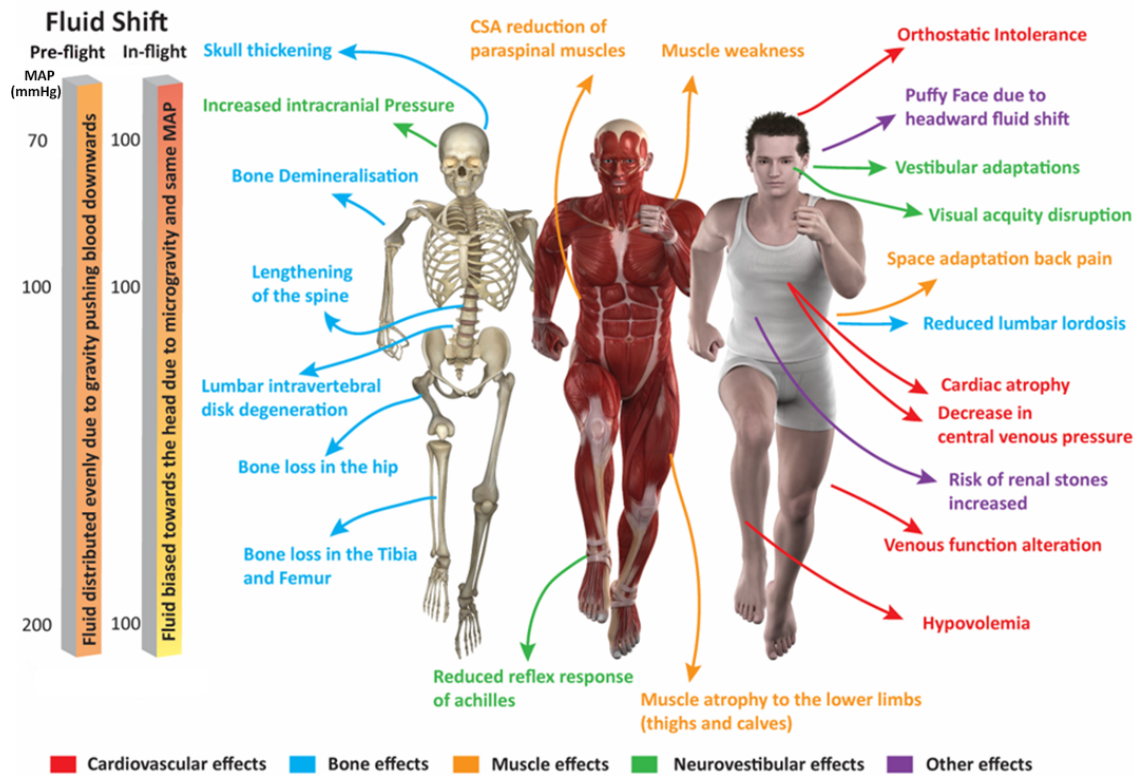

**Figure 1:** Physiological adaptations due to microgravity in outer space. Parts of the figure have been sourced and modified from [7] and Shutterstock.com (license - 182406275)

## 2.1 Weightlessness/Microgravity

Weightlessness and microgravity are used interchangeably. However, microgravity refers to environments where gravitational forces are present but significantly reduced, and weightlessness refers to the sensation or condition of not feeling any weight. In orbit, the gravitational force continues to act on an object, such as a satellite or the ISS. However, as it travels at very high speeds, forces from travelling forwards equals that of the falling, and the object ends up staying more or less at the same height in the curved orbit. This results in a state of constant freefall, causing the inhabitants to experience weightlessness [8, 9].

On Earth, our bodies have adapted to gravity, which ensures the uniform distribution of bodily fluids crucial for homeostasis and mean arterial pressure (MAP) regulation [3]. Microgravity, however, induces a cephalic fluid shift, removing hydrostatic pressure from tissues, muscles, and bones, leading to deconditioning (as graphically illustrated in Fig. 1). The musculoskeletal system experiences one of the most pronounced effects of microgravity. Anti-gravity muscles, such as the soleus, gastrocnemius, quadriceps femoris, spinal postural muscles, and leg extensors, are vital for posture, balance, and movement on Earth. Microgravity mechanically unloads these muscles, inhibiting mechanotransduction signalling for protein synthesis [10]. Consequently, muscle atrophy, particularly in anti-gravity muscles, and bone resorption occur. Upon returning to Earth, astronauts must undertake intense rehabilitation to re-adapt to the Earth's environment, counter the bone resorption, and muscle atrophy experienced during spaceflight.

## 2.2 Other Critical Stressors

### 2.2.1 Radiation

Ionising space radiation, including X-rays, Gamma Rays, and other high-frequency waves, presents significant cancerous risks to humans during space travel, requiring cautious planning and safety countermeasures, especially for interplanetary or moon expeditions [11–13]. Earth's magnetic field partially protects the ISS [14], but it is still susceptible to geomagnetically trapped radiation, *galactic cosmic rays* (GCR), and solar flares (Fig. 2). Acute and chronic radiation risks in humans may lead to short- and long-term health effects, such as cancer or *central nervous system* (CNS) disruptions [15].

The average annual radiation exposure to humans in the US is around 6.2 mSv. During a 6-month ISS mission, crew members experience exposure between 50 to 100 mSv, with higher levels during extravehicular activities (EVA) [16]. It is predicted that a mission to Mars may result in over 1000 times the annual exposure on Earth. Above 1 Sv of ionising radiation can cause acute symptoms and potentially have fatal consequences [17]. Radiation exposure increases free radicals in the body, resulting in increased oxidative stress leading to degenerative and physiological disorders. Animal studies show increased cancellous bone loss and higher osteoclasts (bone-resorbing cells) [17, 18]. Manifestations depend on several factors such as mission parameters, shielding, individual sensitivity, and absorbed dose [19, 20]. Therefore, management and countermeasures are crucial for maintaining physiological functioning and integrity during deeper and longer space missions.

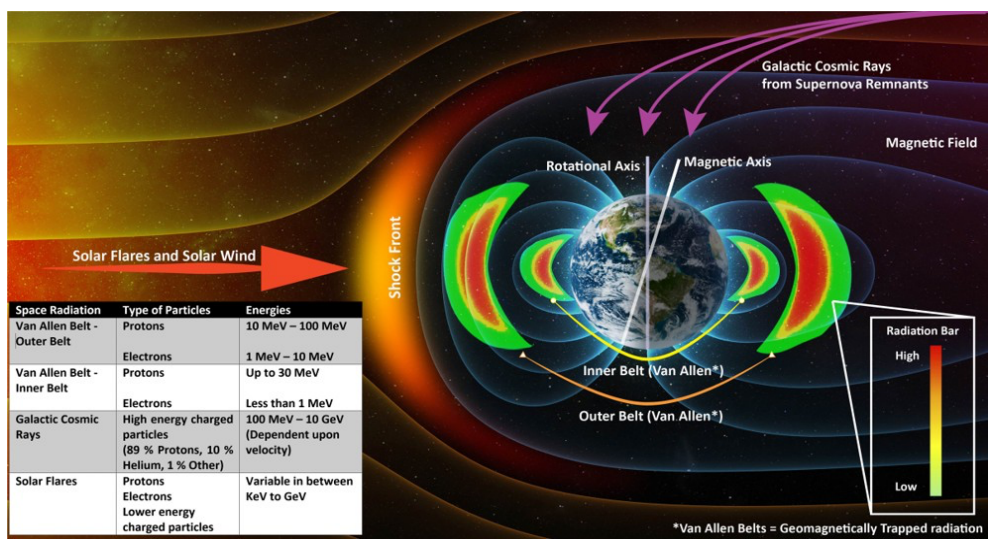

**Figure 2:** Space radiation with their respective energies. Parts of the figure have been sourced and modified from Shutterstock.com (license - 121554235.)

### 2.2.2 Psychosocial Stressors

Cognition, mental well-being, and mindfulness can be impacted while in prolonged space missions due to psychological stress [6, 21, 22]. Individual responses to stress vary, making cognitive appraisal a determinant of psychological stress and its consequences [23]. Additionally, physical stressors can also contribute to psychological stress. Conversely, psychological stress can result in musculoskeletal, cardiovascular, and neurological manifestations. Recent research has linked stress to increased bone resorption and muscle protein degradation [1, 24-27]. Monitoring and managing psychological stress objectively are crucial for mission success, necessitating the development of novel sensors and countermeasures.

## 3. Musculoskeletal Adaptations in Space

This section presents the effects of the different stressors on the musculoskeletal system from the perspective of physiological adaptations and injuries.

### 3.1 Human Locomotion and Movement

Human locomotion on Earth evolved and adapted under Earth's gravity [28]. In space, the reduced ground reaction forces (GRF) result in decreased muscle force, affecting normal human locomotion [29]. Stride length and walking speed are influenced by gravity. Froude number (which is the ratio of inertial to gravitational forces or kinetic to potential energy = $Fr = v^2/gL$, where v is velocity, g is the gravitational acceleration, and L is leg length) of 0.25 and 0.50 correspond to optimal walking speed and walk-to-run transition on Earth, respectively [30-35]. In reduced gravity, optimal walking speed and walk-to-run transition decrease (Fig.3) [30]. Humans at lower gravity (e.g. on Mars) may prefer running at lower velocities. On the moon, astronauts tend to hop rather than walk. This strategy minimises metabolic costs and increases efficiency and stability [35-38].

GRFs, influenced by gravitational acceleration, are lower in partial gravity or microgravity. Schaffner et al. [39] and Genc et al. [40] reported GRF decreases under reduced load and increases at higher speeds, but was significantly lower than on Earth (46% and 25%, respectively). Reduction in bone mineral density (BMD) may also have contributed to reduced GRF. Exercise devices/protocols in the ISS are designed to prevent bone density loss, but usually they do not induce forces comparable to those on Earth [41, 42]. Proposed countermeasures, including bungee ropes, have yet to provide regular forces equivalent to Earth's 1G (= 9.8m/s2) gravitational pull [43, 44].

A different aspect of locomotion, stability, depends on head and gaze coordination with sensory and motor information from the central nervous system (CNS) [45]. In microgravity, the vestibular-ocular system unloading affects balance and gaze control, leading to disorientation [46, 47]. Post-spaceflight, astronauts face an increased risk of tipping [48, 49]. Disturbed gaze control, locomotion, and posture, along with muscle atrophy and bone resorption could result in long-term manifestations or injuries [50]. Locomotion and movement during EVA or IVA (intravehicular activities) have resulted in musculoskeletal injuries and traumas to extremities, back, and neck [51]. Further research is necessary to design effective countermeasures for human locomotion and movement in space, benefiting astronauts and, eventually, those suffering similar disorders or traumatic injuries on Earth.

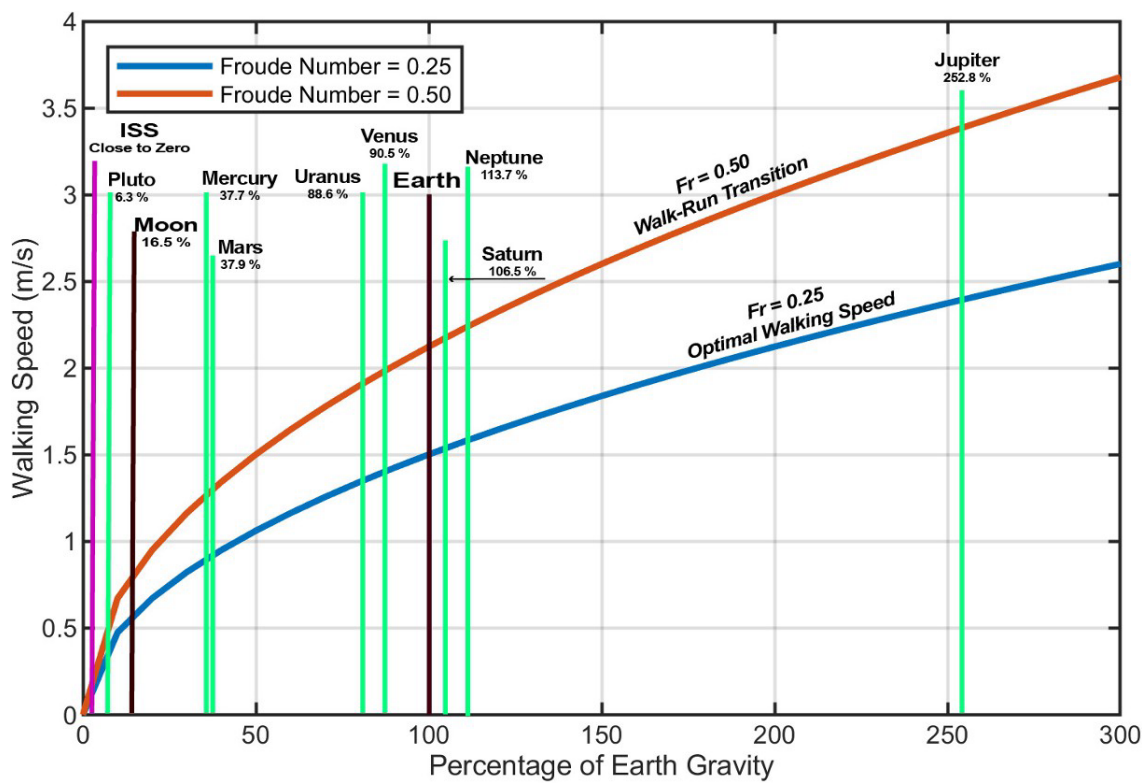

**Figure 3:** Simulated different optimal walking speeds as a function of gravity. Figure generated based on data from [30].

### 3.2 Spine and Back

On average, in microgravity, the human spine extends by 4-7cm [4]. Spinal lengthening is believed to provoke tension in the dorsal nerve roots in the lumbar spine [52], which provokes back pain, referred to as *space adaptation back* pain (SABP). Wing et al. reported that 14 out of 19 astronauts experienced moderate lower back pain with an increase in spine height of 2.1 cm within the first three days of spaceflight [53]. These findings were confirmed by a head-down tilt bed-rest (HDBR) experiment [54]. SABP impairs astronauts' mood, concentration, and performance. Adopting a knee-to-chest position, or 'fetal tuck,' has been found to alleviate back pain [56–58]. Back pain relief has also been observed following in-space exercises and analgesic medications [59]. Astronauts also have four times higher risk of herniated nucleus pulposus (HNP) compared to non-astronauts (data comparison between 983 healthy non-astronauts and 321 astronauts after spaceflight acquired from the Longitudinal Study of Astronaut Health Database) [60].

Magnetic Resonance Imaging (MRI) and ultrasound-based studies (performed pre-flight, immediately and 30 days post-flight) reported reduced spinal *muscle cross-section area* (CSA), decreased lumbar lordosis, reduced bone mass, muscle weakness and para-spinal muscle reduction/atrophy, increased spine stiffness, widespread spinal microfractures and inter-vertebral disc (IVD) degeneration [55-69]. Astronauts in the studies [67-69] underwent resistive exercise using ARED (which provides axial compressive forces to the lumbar region), and this may also have affected the results observed in the lumbar compared to the cervical region, although muscle atrophy was seen in the lumbar IVDs as well.

**3.3 Muscle and Bone Atrophy**

In 1962, Mohler first raised concerns about muscle atrophy in astronauts [70]. NASA's Skylab experiments in 1975 recorded muscle atrophy in crew members and discussed potential countermeasures to mitigate the same [71, 72]. Biostereometric measurements conducted on lower-limb (gastrocnemius and soleus) and upper-limb (biceps brachii and brachioradialis) muscles revealed non-significant muscle loss in the arms but significant muscle loss in the trunk and lower limbs [71, 73-75]. Calves experienced more atrophy than thighs [76–78], with reduced reflex response in the Achilles tendon reported in Skylab 3 and 4 missions [79]. The soleus muscle was most affected by atrophy, with muscle loss observed even in short space missions [76, 77]. Ground models that mimic microgravity in space, such as bed rest, limb immobilisation, and water immersion, also reported muscle atrophy and bone resorption, although at lower rates than in space [80]. Muscle atrophy by soleus and gastrocnemius fibre type is illustrated in Fig. 4 [81].

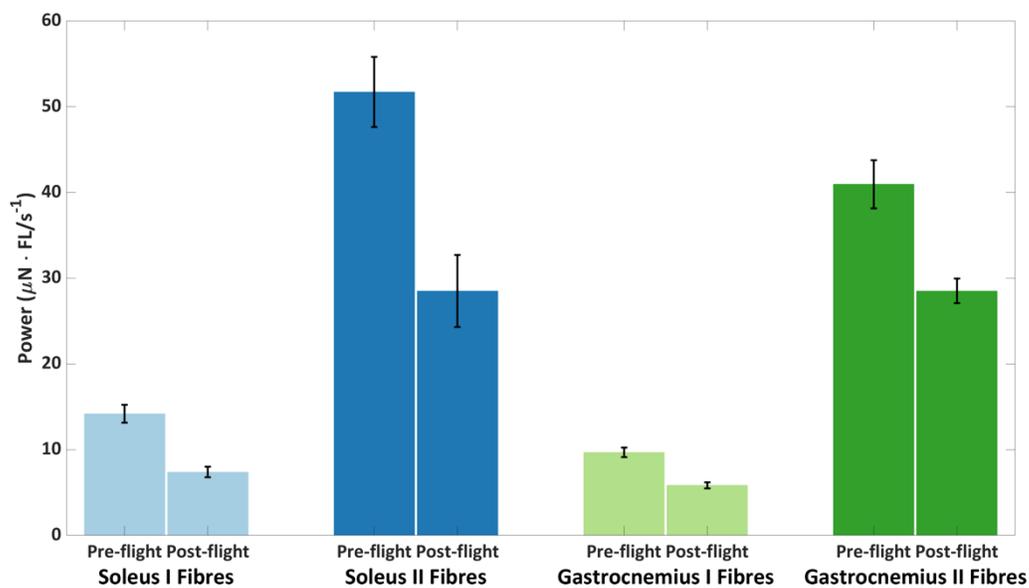

**Figure 4:** Muscle atrophy by fibre type of soleus and gastrocnemius measured at maximal contraction force in pre-flight and post-flight of 10 astronauts. Data obtained from [81]

Bone resorption occurs at different rates in various body regions during space missions [80]. The bone resorption rate for the spine, neck, trochanter, and pelvis is 1.06%, 1.15%, 1.56%, and 1.35% per month, respectively, despite performing resistive exercises and using a treadmill during their stay [82]. Conversely, bone loss is not observed at significant levels in the arm (0.04% per month) and is limited to 0.80-0.90% and 1.2-1.5% per month in the total lumbar spine and the hip, respectively [83]. Multiple studies have supported these findings, including a total of 14% bone resorption that occurred in the proximal femur bone during four to six months of Mir missions [84–86]. LeBlanc et al. [82] reported that the lean arm tissue did not undergo any atrophy, whereas the lean leg tissue underwent atrophy at a rate of 1.00% per month. The bone recovery post-flight was reported to be slow, with approximately 2-3 years needed to regain pre-flight levels, and raises serious concerns of increased risk of osteopenia and fractures during extended space missions [29, 87].

### 3.4 Injuries to the Musculoskeletal System

While the previous subsections discussed the different adaptations the human body could undergo, injuries also are a prevalent reality. Musculoskeletal injuries during space missions (Fig. 5) primarily affect hands, back, and shoulders due to repetitive activities, modified locomotion, restrictive clothing, and microgravity adaptations [51, 88,89]. The restrictive EVA suit, known as the *Extravehicular Mobility Unit* (EMU), while critical for spacewalks and EVA training, also poses challenges to astronaut comfort and safety [88, 90, 91]. The EMU consists of multiple layers, starting with the liquid cooling and ventilation garment (LCVG) that maintains thermal balance through water circulation. This is followed by a rigid fibreglass component called the hard upper torso (HUT) provides structural attachment points [90].

In the vacuum of space, the pressurised EMU suit hinders movement, causing discomfort, fatigue, and skin, muscle, and joint injuries [88]. Poor EMU fit can exacerbate musculoskeletal disorders like microgravity-induced lower back pain [92]. With most activities relying on the upper extremity, overuse and repetition injuries commonly affect the hands and shoulders, as well as the feet [90]. Approximately 34.2% of injuries occur in the hands and 10.7% in the shoulders. Shoulder injuries mainly result from contact and strain at the HUT attachment points [93].

During EVA, fingertip and fingernail injuries are prevalent due to pressurised glove usage, significantly affecting hand strength, dexterity, and comfort [94]. Persistent exposure leads to subungual haematoma (redness), fingernail pain, and onycholysis, potentially escalating to secondary infections [88]. These infections pose a significant challenge, given the decreased efficacy of medication in space and the potential for bacteria to develop resistance [88]. Other issues include fingertip abrasions, frostbite, neuropathies, dislocations, subungual hematomas, and muscle stress [88, 95]. Increased moisture in the glove and reduced blood flow to the fingernail bed also contribute to these problems [88, 95]. An exhaustive analysis of spacesuit glove-induced hand trauma can be found in [96].

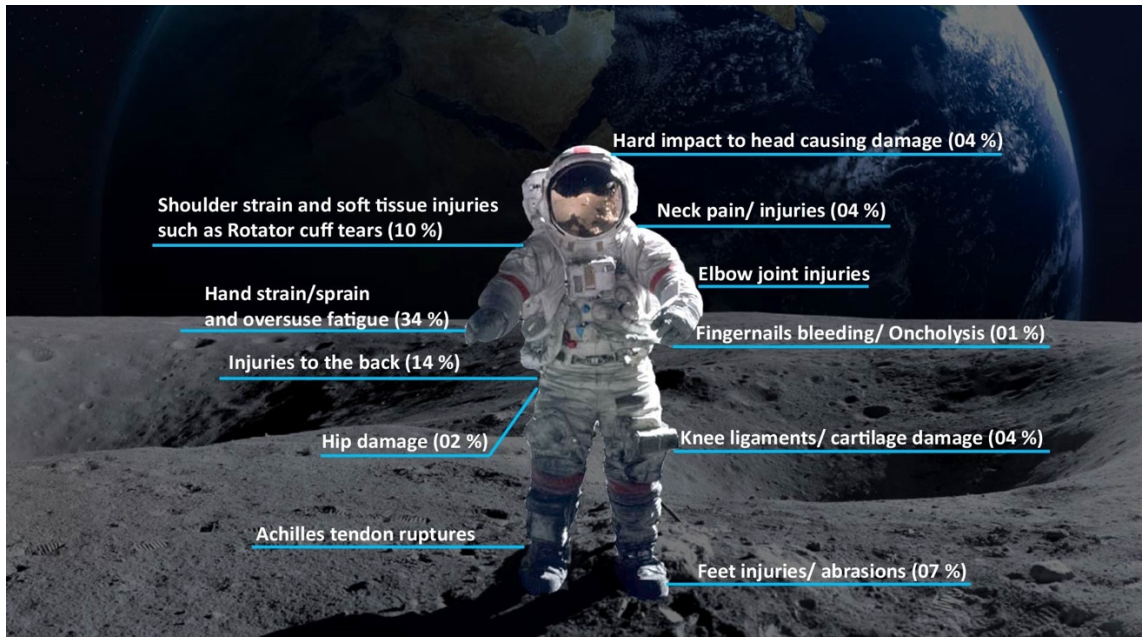

**Figure 5:** Physiological injuries and adaptations during/after spaceflights. Injury data obtained from [93]. Parts of the figure have been sourced and modified from Shutterstock.com (license - 360211214)

Back injuries are the second most common injuries after those of the hands, followed by injuries to shoulders, feet, arms, and neck, due to crew activities and in-flight exercises [51]. The ISS has a higher incidence of back injuries than other spaceflight missions, such as Mir or Shuttle, which is attributed to the introduction of exercise devices, such as the *interim Resistive Exercise Device* (iRED). While these devices aim to reduce microgravity-related physiological adaptations, they have also increased minor physical injuries, primarily strains and sprains, with contusions being less common [51]. Although fractures are rare in flight, bone resorption during spaceflight raises post-flight fracture risk. Consequently, enhanced management protocols are needed for astronauts to safely exercise and prevent in-flight and post-flight injuries.

## 4. Countermeasures

Aerobic and resistive exercises have been some of the earliest countermeasures proposed for the adverse effects of microgravity. These aim to load the lower extremities. Different space agencies have been investigating additional countermeasures, ranging from specific exercise devices to artificial gravity (AG), pharmacological interventions, and nutrition. Through the Human Research Program [97], NASA has been investigating solutions combining traditional solutions (e.g., exercise, nutrition, and pharmacological interventions) and more unconventional solutions (e.g. AG, neuromuscular electrical stimulation (NMES) and vibration). Research has shown that efficacy reduced from exercise to nutrition, pharmacological interventions, AG, NMES/FES, and vibration [97, 98]. Resistive/aerobic exercise remains the most effective countermeasure, with other methods contributing to improved effectiveness [98]. For nutrition, the impact of eucaloric v/s hypocaloric v/s hypercaloric intake, protein intake management and supplementation have been investigated, and pharmacological interventions such as bone-resorptive medication (bisphosphonates) and low-dose testosterone doping have also been tested [98]. However, in this review, we will focus primarily on physical/exercise-based countermeasures. It should be noted that while literature is available discussing the physiological implications of using the different exercise countermeasures, very limited information is available about the engineering specifications of the different technologies discussed next. The available information is mostly from public dissemination information on space agency websites and as part of background sections in some literature.

### 4.1 Exercise Countermeasures on the ISS and other Space Missions

Because of the reasons discussed above, astronauts follow a strict regimen of exercise, diet, and pharmacological supplementation to minimise bone loss/muscle atrophy. While exercise devices started out with simple resistance bands (Fig.6a), bungee cords and treadmill-like devices (Fig.6b), devices are now much more sophisticated. Astronauts now spend about two hours daily on specialised equipment such as treadmills like the *Treadmill with Vibration Isolation Systems* (TVIS) [99], T2, and *Combined Operational Load Bearing External Resistance Treadmills* (COLBERT) (Fig.6f), stationary bikes such as the *Cycle Ergometer with Vibration Isolation System* (CEVIS) [100] (Fig.6c&g), and resistive exercise equipment (ARED (Fig.6e), iRED (Fig.6d)) [101] to counter the effects of skeletal muscle unloading experienced in space [102]. Newer resistive and aerobic exercise devices, such as the *Functional Re-adaptive Exercise Device* (FRED) [103] (Fig.6h) and SoniFRED [104], are currently under development and testing. However, current countermeasures are not fully effective, and new or improved solutions are needed [4, 105].

Treadmill use in Skylab 4 reduced leg atrophy, with contribution from the higher nutrition intake as well [78], but bicycle ergometer and exerciser protocols did not yield significant benefits in preventing muscle atrophy. The iRED platform which provides resistance of up to 125 kg via elastic cords and was introduced in the ISS in 2000. However, experimental results proved it could not provide sufficient/constant resistance to prevent lower leg muscle atrophy or bone resorption. Therefore, in 2009, the ARED was introduced. It utilises vacuum cylinders along with flywheels so that astronauts can perform leg exercises, such as squats and deadlifts. ARED has now fully replaced iRED as it provides significantly greater resistance of around 275 kg, though crew members can only exercise in one axis (1 degree-of-freedom movements). Smith et al. [106] reported that, upon returning to Earth, astronauts exercising with ARED in space had a higher proportion of lean mass and lower fat mass when compared to astronauts performing exercise with iRED. Although ARED is more effective, it is time-consuming and has been reported to hinder visual acuity. A combination of ARED and pharmaceutical interventions (e.g., bisphosphonates) showed promise in preventing bone loss, but their long-term side effects in space are unknown [107, 108]. As mentioned previously, some sprains and strains were recorded as a result of regular usage of the ARED. It should also be noted that balanced mechanical systems such as the ARED and others as opposed to free weights, do not allow for simultaneous training of stabiliser muscles and, hence, can be less effective, but would also result in lesser injuries.

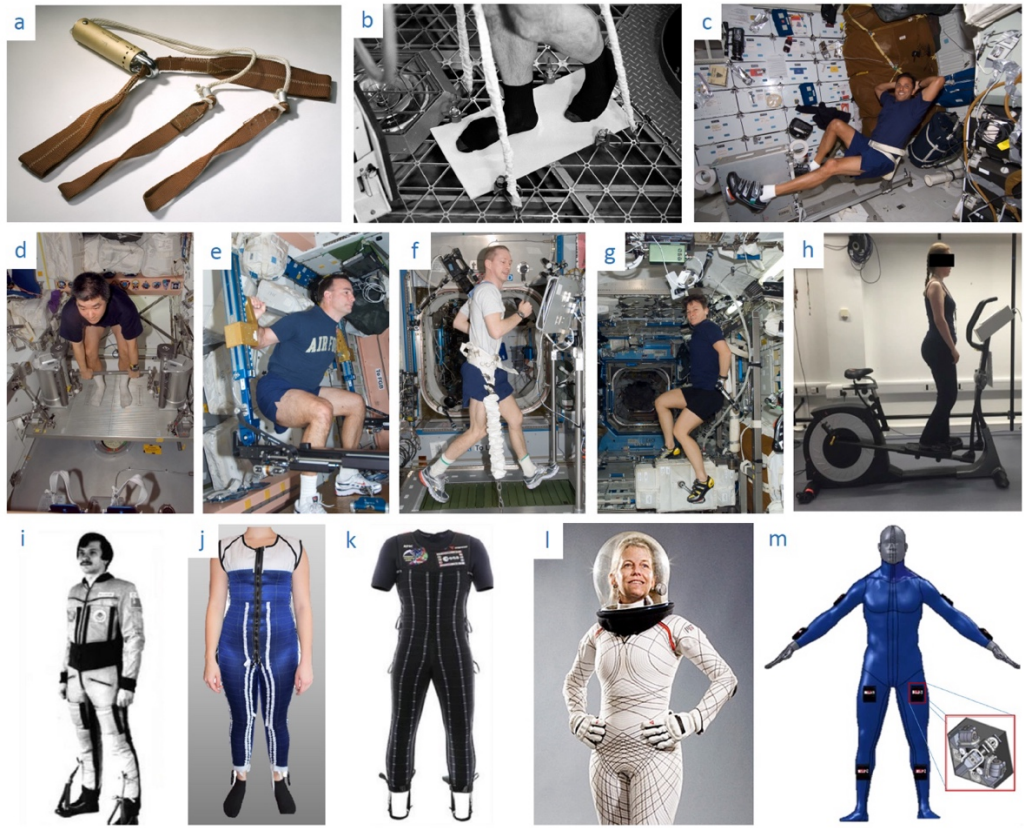

**Figure 6:** Overview and evolution of deployed and under-development countermeasures. **Top Row: a).** the exerciser used on the Apollo 11 mission (Photo by Eric F. Long, Smithsonian National Air and Space Museum [NASM 2009-4775] ©Smithsonian; **b).** a Teflon-coated treadmill-like device used during Skylab 4 (Photo Credit: NASA); **c).** Cycle ergometer device used onboard the Discovery space shuttle ©NASA; **Middle Row: d).** iRED device used on Expedition 16 flight ©NASA; **e).** ARED device used during the space shuttle missions ©ESA/NASA **f).** COLBERT treadmill device used on Expedition 21 flight ©ESA/NASA **g).** CEVIS device used onboard the ISS ©ESA/NASA **h).** the under-development FRED device [144]; **Bottom Row: i).** the Pingvin/Penguin (Adeli) countermeasure suit [115]; **j & k).** different versions of the Gravity Loading Countermeasure (GLCS) Suit [116]; **l).** the Biosuit, a mechanical counterpressure suit [117]; **m).** the Variable Vector Countermeasure Suit (V2Suit) [118].

The FRED (Fig.6h) has been developed to provide lumbar-pelvic reconditioning (with a focus on the *lumbar multifidus* (LM) and *transversus abdominis* (TrA) muscles) and reduce symptoms of SABP [114]. A combined intramuscular and surface electromyography (sEMG) study by Weber et al. [109] using FRED found sustained activation of the TrA and LM muscles, endorsing its use as a countermeasure. Despite promising results in reducing or preventing LM and TrA deconditioning, FRED has not been deployed in the ISS yet. To counter SABP, researchers have also employed virtual reality to help users achieve correct orientation and posture [110] and to provide vestibular information for restoring proprioception in microgravity.

Resistive exercises have been shown to help prevent bone resorption in some areas but not in others, such as the lumbar spine and hip [84, 87, 111]. Rittweger et al. [111] found that resistive exercises during HDBR prevented bone resorption in the tibia but not in the radius, lumbar spine, or hip. Results from long-duration spaceflight missions showed significant bone resorption in the lumbar spine and hip [84, 87]. In a study using flywheel resistive exercise during HDBR, calf muscle atrophy and bone resorption were partially reduced [111]. This resistive exercise device enabled a gravity-independent workout and was found to increase muscle volume and strength.

*High-intensity interval training* (HIIT), a popular protocol for alternating periods of high and low-intensity exercise, also proved to be effective as a countermeasure against microgravity [112]. This protocol improves neuro/cardio/muscular/skeletal-fitness by increasing muscle peak power, lean muscle mass, and lung capacity. The protocol has been used since the earliest space missions, but the choice of exercises, duration, and effectiveness are now being optimised. The intensity, frequency, training type (aerobic or resistive) and loading are essential considerations when evaluating countermeasure technologies and protocols [113].

## 4.2 Artificial Gravity

*Artificial Gravity* (AG) has also been proposed as a potential solution to mitigate space-induced microgravity effects [133, 134]. The Japanese Space Agency (JAXA) implemented an AG system for mice and reported some success. However, implementing a successful system for humans remains challenging due to unknown optimal parameters and limited studies supporting its efficacy [135–137]. Also, spinning an entire spacecraft is cost-ineffective and presents safety hazards, so the focus has been on short-radius centrifuge systems [133]. Creating AG equivalent to 1G requires significant angular velocity (approximately 30RPM), and could cause transitional adaptation consequences, such as vestibular disruptions [138].

## 4.3 Neuromuscular/Functional Electrical Stimulation

*Functional Electrical Stimulation* (FES) or *Neuromuscular Electrical Stimulation* (NMES) is a promising intervention that uses electrical impulses to activate muscle contractions. It provides a targeted approach to maintaining muscle mass and effectively counteracting muscle atrophy in microgravity environments. Mayr et al. [124] used an EMG-NMES system called MYOSTIM-FES on four muscle groups (quadriceps, hamstrings, tibialis, and triceps surae) to apply electrical stimulation (generating 20% of maximum voluntary force) for six hours daily. Although detailed results are unavailable, they found a 5% reduction in atrophy and increase in Type 1 and Type 2 fibres. The technology was non-intrusive and user-friendly. Duvoisin et al. [125] reported improved muscle volume, mass, and CSA of twitch fibres in astronauts, using the same technology.

A recent study [126] using NMES/FES on the triceps brachii reported increased muscle mass but no significant change in strength. On-ground studies on the quadriceps group also echoed these findings [127, 128]. It is hypothesised that NMES may attenuate myostatin pathways that preserve mass but do not restore signalling pathways for strength [127]. FES/NMES exploration as a countermeasure is recent and the optimal electrical stimulation parameters are still under investigation. In the late 90s, NASA developed the StimMaster FES Ergometer [129] and the Percutaneous Electrical Muscle Stimulator II [130, 131], used in the ISS's human research facility. NMES/FES has the advantage of providing selective muscle recovery and activating type II muscles at lower forces, making it a promising complementary countermeasure [132]. However, its long-term effectiveness and safety must be assessed for extended space missions as FES has been known to suffer from limitations such as providing reliably consistent stimulation and is also known to induce fatigue.

**4.4 Limitation of Current Exercise Countermeasures**

Although physical exercise reduces musculoskeletal deconditioning, excessive exercise can increase the probability of injury and also increase free radicals and cause oxidative stress. Hence, other alternative countermeasures in addition to exercise are also being researched. NASA and other space agencies are creating spaceships that are significantly smaller than the ISS, making the currently bulky and extensive exercise countermeasures infeasible. For instance, the Lunar Gateway spaceship [119], being developed as part of the Artemis programme, is only 12.5 % of the ISS's size. Similarly, the *Tiangong Space Station* (TSS) [120], Axios Space Station [121], and Bigelow Aerospace B330 [122] are all future *Low Earth Orbit* (LEO) space stations that are significantly smaller than the ISS. Most of these space stations and spaceships have been designed for missions in the range of a few weeks for 4-6 astronauts. Secondly, while costs to carry loads to space have come down significantly since the first missions which were over $41,000/kg during the Space Shuttle, it is still high at >$1700/kg even accounting for the economics of reusable rockets. This would make most platforms such as the TVIS, CEVIS, ARED and others to be unsuitable for the stations. Apart from concerns of weight, bulk and extent of current countermeasures, there are concerns regarding compatible materials, electronics insulated from EMI and radiation, vibration insulation, thermal sensitivity, leveraging existing power sources whether pneumatic or electric, etc. Hence, research for novel bespoke/individual-specific countermeasures is crucial for the upcoming age of interplanetary space travel [123].

# 5 The Opportunity for Wearable Technologies as Countermeasures

**5.1 Experimental Wearable Technologies on Space Missions**

Wearable systems such as the Pingvin (Fig. 6i) and *Gravity Loading Countermeasure Suit* (GLCS) (Fig. 6j&k) suits have been investigated to mitigate the effects of microgravity since the 1990s [115,116,139,140]. These suits, designed to emulate gravity passively via strategically positioned elastic bands or weaves, were tested on the Mir and ISS, respectively. The Pingvin suit, the predecessor to the GLCS, delivered a 0.5G load without exercise but was criticised for discomfort and inadequate thermal conductivity.

The GLCS, drawing lessons from the Pingvin suit, incorporated bidirectional elastic weaves to enable variable axial loading through tension application [141]. On average, the GLCS managed to impose a 0.7G load. However, it was reported to be quite restrictive and slowed locomotion. Despite these drawbacks, the GLCS demonstrated potential benefits, including improved ventilatory response, reduced perceived workload in microgravity, and decreased spinal elongation, suggesting a potential role in mitigating back pain [141, 142] Another technology being developed by the same laboratory is the Biosuit, a mechanical counterpressure suit leveraging the body's strain fields and the concept of lines of non-extension. This design was proposed as an alternative to pressurised spacesuits, and consequently reduce the discomfort and potential injuries suffered by astronauts [117].

More recent research has focused on the development of the Variable Vector countermeasure suit (V2Suit) (Fig.6m) [143,144]. The V2Suit utilises wearable modules comprising an *Inertial Measurement Unit* (IMU) (Fig. 8h) and a *control moment gyroscope* (CMG) to deliver dynamic resistance to different body segments. The IMU tracks orientation, position, and motion, while the CMG generates torque and resistance [144]. Integration of V2Suit with GLCS could enhance dynamic loading to counter muscle atrophy, although the effectiveness hinges on the algorithm's robustness and the accuracy of the IMU's estimations [143].

As part of its *Game Changing Development* (GCD) program [145], NASA co-developed several exoskeleton/exosuit technologies for in-/post-flight exercise, rehabilitation, and assistance applications. These wearable systems could function as a portable gym for astronauts, providing constant muscle loading and potentially replicating gravity, which could supplement/replace traditional gym sessions. In the space-constrained ISS environment, where time and productivity are crucial, wearable devices could offer full-body monitoring, apply specific resistance profiles for training, and function as both assistive and resistive devices through a simple change in control strategy. NASA's co-developed X1 lower-limb exoskeleton (Fig.7i) with the Florida Institute for *Human and Machine Cognition* (IHMC) serves as a resistive exercise device in space and for terrestrial assistive applications [146]. The *European Space Agency* (ESA) and its telerobotics laboratory also developed several systems such as the EXARM, X-ARM-II, SAM, and the ESA exoskeleton (Fig.7j) [147]. However, these exoskeletons were developed for the applications of telepresence and haptic feedback, and not specifically for assistance or rehabilitation. NASA has also embarked on developing softer systems, including the Robo-Glove (Fig.7k) [148], a spin-off from the Robonaut 2 project developed in collaboration with *General Motors* (GM) to aid in physically intensive and repetitive tasks. The Armstrong system (Fig.7l), co-developed with Rice University, aims at shoulder augmentation and rehabilitation [149]. Similarly, the Japanese and Russian space agencies have also investigated and invested in exoskeleton technologies. As most of these systems were developed at space agencies and not as part of academic research, detailed descriptions are not always available for a concrete analysis. Details of research-based and some commercial wearable robotic and sensing systems, will be discussed in the next section to further present their potential as alternatives to current countermeasures.

## 5.2 Wearable Robotics in Terrestrial Applications

The potential of wearable robotics and sensing technologies to monitor musculoskeletal health and to provide an active countermeasure through dynamic muscle loading or assistance could be game-changing. Within the context of space-based applications, wearable robotic technologies could have applications ranging from providing dynamic muscle loading and mimicking a sense of gravity to providing active assistance during physically strenuous IVA/EVA tasks. Based on the above, exoskeleton technologies can be broadly divided based on the magnitude of force transmission from the robot to the wearer.

### 5.2.1 Medium-to-High Force Applications

Astronauts performing EVA are required to wear the EMU suit that is made up of multiple layers, concluding with the HUT. In the vacuum of space, this pressurised suit can severely hinder movement, requiring the astronaut to fight both against the suit and the physical activity for hours together, causing significant discomfort, fatigue, and potentially skin, muscle, and joint injuries. Furthermore, EMU fit misalignment can exacerbate injuries and musculoskeletal disorders like microgravity-induced lower back pain. With most activities relying on the extremities, we noted that the overuse and repetition injuries commonly affect the hands, feet, and shoulders. For assisting with these medium-to-high force applications, a rigid-bodied exoskeleton technology integrated into the suit is a potential future solution for the astronaut. In addition to assisting during IVA/EVA, these compact exoskeletons could also provide constant resistance during different activities, mimicking the effect/resistance of Earth's gravity and/or providing constant loading and acting as a wearable gym for the astronaut.

Amongst exoskeleton technologies, rigid-bodied exoskeletons have been the dominant design architecture, providing a rigid frame and facilitating high force/torque transmission (Fig. 7a-d) [150–153]. These systems can be either actively (Fig. 7a-b) or passively (Fig. 7c-d) actuated. For active systems based on more sophisticated controller designs, the same hardware platform could, in the future, achieve a substantial increase in performance. Ideal applications include heavy industry, military, rehabilitation for patients with spasticity or other high force/torque requirements, and extravehicular space applications. A number of these systems have obtained regulatory approval and have now been commercialised [154–156]. In terrestrial applications, lower-body systems (Fig.7a-b) are more prevalent than upper-body systems (Fig.7c-d), with few systems achieving full portability [155, 157–160]. The rigid-body design archetype allows for high forces to be grounded through the frame of the exoskeleton (in place of the body, which is the case in soft exoskeletons) allowing for higher force transmission.

However, rigid-bodied exoskeletons have disadvantages, such as being heavy and extensive, requiring high-torque actuators, and large power sources [161, 162]. Lightweight systems using materials like carbon fibre can improve wearability but may still impede range-of-motion for basic *activities of daily living* (ADL) [153, 163–165]. The sub-optimal mechanics of engineered rigid-bodied systems compared to the complex biomechanics of the human body could also result in them being restrictive or constraining movements and being a potential source of discomfort, pain, or injury [152, 166–168]. Systems designed with self-aligning [169] or self-adapting [170] mechanisms may alleviate this problem but would come at the cost of increased size and weight [171–174]. Rigid exoskeleton design is an exercise in optimisation of the size and weight of actuators, power sources and frames which can be part of a vicious circle, and the design of heavy/bulky systems has resulted in the discontinuation of military projects such as HULC and XOS [150, 152, 175]. However, advances in power storage, materials, actuation, and sensing technologies are making rigid-bodied exoskeletons increasingly realistic for real-world applications.

In space, while microgravity leads to many maladaptations in the human body, it provides an advantage for the adoption of rigid exoskeletons as weight is no longer a limitation. However, bulk/size/inertia of the device continues to be so. This allows for more sophisticated designs, more powerful actuators, or heavier power sources (batteries or compressed fluid storage). Additionally, microgravity allows for actuators to not have to fight against the effects of gravity but only inertia, friction, and other gravity-agnostic forces, breaking the vicious cycle that rigid-bodied systems suffer in terrestrial applications. This advantage in space could make rigid-bodied exoskeletons an implementation of choice for assistance with challenging and labour-intensive EVA activities.

Traditionally, rigid-bodied exoskeletons have used electricity or pneumatics/hydraulics as energy sources [180, 181]. Electromagnetic actuators like motors are prevalent in exoskeletons due to their availability, reliability, ease of installation, operation, and control [182]. They can have a comparable power-to-weight ratio to pneumatic and hydraulic actuators (if the weight of the compressor tanks etc. is included) [180, 183]. Examples of systems using motors include the Indego exoskeleton (Fig. 7a), the Stuttgart Exo-Jacket (Fig.7b), HAL Single Joint Elbow [154], and Hand of Hope [184], with various transmission systems both using direct-drive [185], and indirectly through tendons [186], linkages [184], and chain-and-sprocket [187] mechanisms. Compared to pneumatics and hydraulics, *brushless direct current* (BLDC) or DC motors suffer from high intrinsic impedance (low compliance/high stiffness), making them potentially unsafe, especially in the case of unexpected events like control system malfunction, power failure or spasmodic events [180, 181]. Using springs or hydraulic pistons as elastic elements in series with motors, *Series Elastic Actuators* (SEAs) reduce intrinsic impedance and provide more compliance but have a narrow functional bandwidth [188–190]. Sophisticated human-in-loop controllers are now being developed to realise individual-specific profiles to maximise assistance [191,192].

Pneumatic actuators can be lightweight, have low impedance, and have a higher power-to-weight ratio (if compressed air storage is not considered), allowing for reasonable forces with a light frame. Systems based on conventional pneumatic systems, such as cylinders/pistons were developed, e.g. for finger movements [193] and wrist pronation/supination [194]. *Pneumatic Artificial Muscles* (PAMs), a different approach using pneumatics inspired by biological muscles, consisting of an internal bladder and braided mesh shell (and when pressurised, the actuator expands its diameter and shortens in length just like biological muscles), have been widely used in rehabilitation and assistance applications in both rigid-bodied [195, 196] and soft-bodied [197, 198] systems. Building up on biomimicry approach, they have been used directly [195] or indirectly through tendons and linkages [199, 200]. They have been used for powering single joints and have been used for full-bodied humanoid systems as well. These actuators have limited displacements but can exert high forces. However, similar to other soft actuators, they suffer from a nonlinear nature which has been well-documented and modelled.

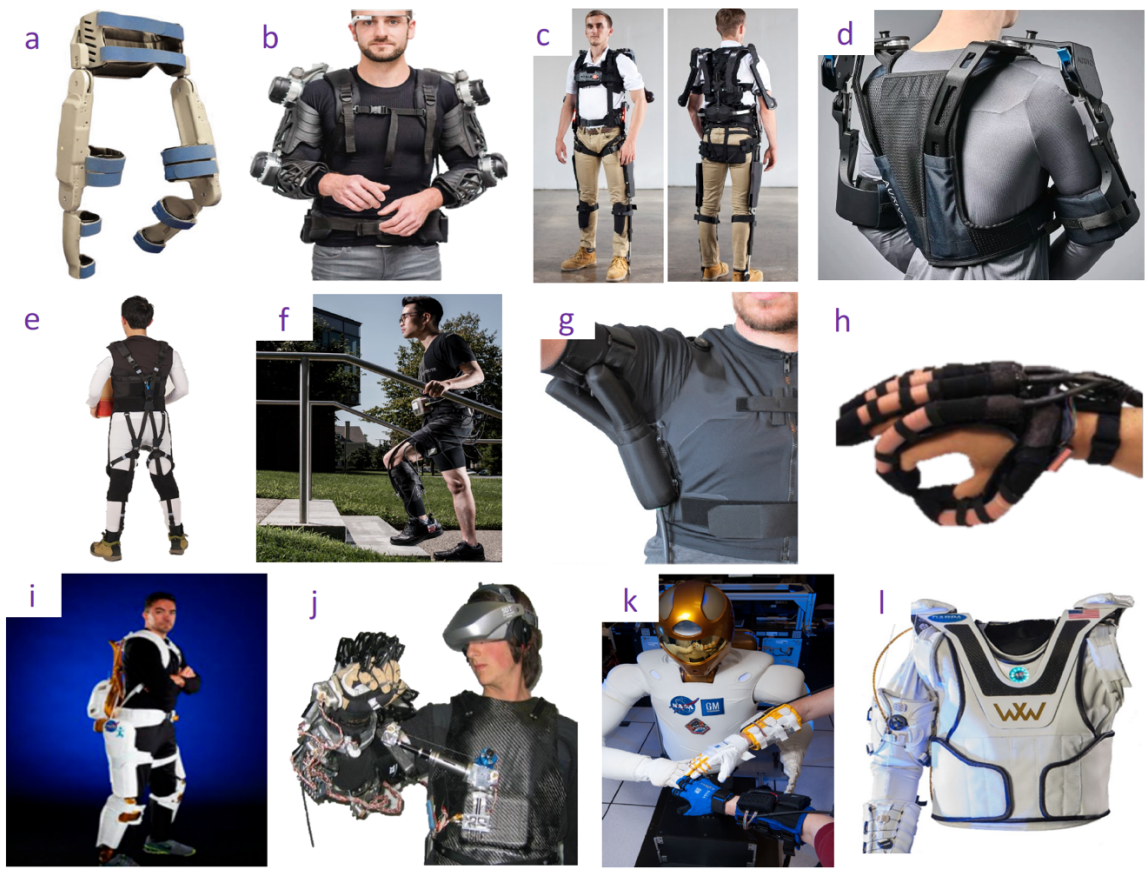

**Figure 7:** An overview of wearable robotic systems developed for terrestrial and space applications. **Top Row: a).** the Vanderbilt powered orthosis, now commercialised as the Indego exoskeleton [186]; **b).** the Stuttgart Exo-Jacket [160]; **c).** the SuitX exoskeleton consisting of the shoulderX, backX and legX components (US Bionics, now acquired by Ottobock) [156] ©Ottobock; **d).** the DeltaSuit by Auxivo ©Auxivo; **Middle Row: e).** a body-powered variable impedance suit to reshape lifting posture [176]; f). a tendon-driven ankle exosuit, now commercialised by ReWalk [177]; **g).** a pneumatic exosuit for restoring arm function [178]; **h).** a fluid- powered soft robotic glove [179]; **Bottom Row: Robots developed for space-related applications: i).** the NASA X1 exoskeleton co-developed with IHMC ©NASA [146]; **j).** a exoskeleton developed at the ESA for telerobotics/telepresence applications ©ESA [147]; **k).** the NASA RoboGlove co-developed with GM ©NASA; **l).** the Armstrong shoulder exosuit co-developed with Rice University ©NASA [149].

Hydraulic actuation, used in a few rigid-bodied exoskeletons such as [183, 201], could also be an alternative option. The actuators offer advantages over pneumatic actuation, such as more force delivery/loading and better efficiency [180, 181]. However, hydraulic actuators are often avoided due to their weight, slow response time, noise, and fluid leakages (and associated environmental concerns) [183]. Additionally, in space, using hydraulics or pneumatics has an additional challenge of managing large temperature variations, and hydraulic fluids could lead to short circuitry and other damage. *Hydraulic Series Elastic Actuators* (SEAs) combining hydraulics and electric motors have also been used in fixed platforms like the NeuroEXOS system [202]. Apart from weight not being a limitation anymore, pneumatic systems can be potential actuation sources as they can be built on cutting-edge knowledge and implementations on spacecraft and space stations and leverage pre-existing resources, including access to compressed gases. Apart from electromechanical, pneumatic and hydraulic, a few systems using shape memory allows, twisted string actuators and some other nascent principles have been developed, however, functional capabilities will be inferior to those discussed above as these systems have much lower force production capabilities and lower dynamic response. However, these and other more novel actuation principles find more promise with soft-bodied systems and these and the concept and latest developments in soft bodied wearable robots are discussed next.

### 5.2.2 Low-to-Medium Force Applications

While the previous subsection discussed medium-to-high force transmission, there is also the opportunity for low-to-medium force applications. In space, this particularly would pertain to technologies that can be worn for extended periods without being obtrusive or causing discomfort. The technology could act as an active wearable countermeasure, providing a sense of partial gravitational loading, and as a wearable gym, causing the wearer to put more effort during ADL, helping maintain muscle tone and preventing atrophy. The GLCS and the Pingvin suit aimed to achieve this passively by trying to return the effects of gravity partially. A second application of the same technology could be at-home rehabilitation post-flight. In terrestrial applications, as weightlessness is no longer an advantage, lightweight, unobtrusive, transparent, compliant, and low-profile systems could prove more advantageous. Technologies based on soft robotics principles could be an ideal use case for this application.

Wearable soft robotic systems are lighter, less restrictive (greater range of motion, lower risk of injuries), and more energy-efficient than their rigid counterparts [153, 179, 203] facilitating assistance for complex joints like the shoulder and hip (Fig. 7e-h). Soft robotics grew as a field as the need for improved physical human-robot interaction and the limitations of conventional robot design in complex environments grew [204]. Researchers drew inspiration from nature, incorporating the compliance of soft tissues [204–206] and embodied intelligence [207, 208] into their designs.

Soft robotic principles become ideal for addressing the assistance of multi-DoF joints with complex biomechanics (such as the wrist, shoulder, and hip), however, current research mostly focuses on single-DoF assistance (Fig. 7e-h) [177, 179, 203, 209–212]. Some systems show simultaneous multi-DoF assistance as a concept [213–216], but only a few have demonstrated such capability [215, 216]. As soft-bodied systems can have a low profile, they can also be concealed and worn under regular clothing, removing psychosocial barriers of advertising weakness/dependency on assistive technology. They also use less expensive and more widely available materials, increasing affordability [152]. The low profile, unobtrusiveness and improved wearability of these systems would also be an advantage inside spacecraft and space stations where real estate is at a premium. Most soft-bodied wearable robotic systems have focussed on hand exoskeletons due to the low force requirements, or on assisting single DoF movements such as the elbow. As we go up the limb, with increased mass/inertia of the limb, the challenges for developing completely soft systems to achieve 100% assistance becomes almost unattainable with few systems capable of multi-DoF shoulder assistance. Similarly, for the lower-limb, most systems focus on assisting the ankle with an aim to assist in propulsion, with systems for the hip being less prevalent or achieving very little assistance.

The biggest drawback of soft-bodied exoskeletons comes from their principal strength: the absence of a rigid external frame. The lack of rigid frame results in limited force/torque transmissibility, absence of force grounding, difficulty to achieve direct drive, and sensor/motor mounting difficulties [154, 162, 217]. Moreover, limited power generation from soft actuation technologies results in lower force/torque generation as well. However, limited force generation may not be a shortcoming in space-based applications, as the actuators do not have to fight against gravity to provide assistance/resistance. This could make space applications an ideal testing ground for soft-bodied systems. However, one of the biggest challenges with a low-profile system that is not directly-driven (regardless of the actuation principle) is that the low-profile nature impacts the moment arm available, consequently limiting the torque applied to the joints. While this is not necessarily a challenge for the fingers or the wrist, developing purely soft systems e.g. the shoulder become challenging. A potential middle-ground built on hybrid systems could provide higher forces. These systems could employ rigid and soft components [198, 218] or materials and mechanisms that stiffen and soften on demand [205]. Another challenge in the design and implementation of soft exoskeletons is the non-linearities arising from the compliance in soft embodiments and the user's body influence the system dynamics, making controller design complicated [198, 199, 207, 219]. Some systems have investigated on-demand variable stiffness to overcome this shortcoming. A review of different stiffening technologies was compiled in [205], while a detailed review of soft robotic suits was compiled by Xiloyannis et al. [220].

Soft exoskeletons have been developed with passive (Fig. 7e) and active (Fig.7f-h) actuation principles. Soft exoskeletons have used a range of novel actuation principles and mechanisms, which will be discussed briefly next. While most systems were developed for terrestrial applications, they provide a holistic overview of the field for further research into space and terrestrial applications. Soft exoskeletons based on passive actuation principles can be efficient from a power/weight perspective, and systems building on top of the GLCS and other concepts leveraging springs, dampers, inertia or other more novel passive elements to apply resistive forces or to force good ergonomics [176] amongst other applications could be attractive especially for space applications. Amongst active actuation principles, DC/BLDC motors have been the most common actuation methods used to control cable/tendon-driven systems. Examples include commercial systems such as the Robotic SEM Glove [152, 221] and the system by Bae et al. (Fig. 7f) [177] commercialised by ReWalk, and other research-based systems [203, 209, 222]. Cable/tendon-driven systems with centrally located actuation packs are more common than direct-drive systems as mounting motors on soft frames is challenging and adds to weight and inertia on the limbs being assisted. Tendon-driven systems offer an easy setup, maintenance, remote actuator placement, controllability, low profile, and higher forces compared to pneumatic and hydraulic soft actuators [203, 223]. Some tendon-driven hand exoskeletons are under-actuated and use only a single actuator for multiple joints [182]. While using cables/tendons, localised forces from the tendons or at attachment points are a concern and require elements for better force distribution. This challenge amplifies as force requirements increase and consequently leads to increased cable tension, such as systems for the shoulder or the hip.

Using pneumatics and hydraulics, soft actuators ranging from single-chamber gloves [224] to more sophisticated elastomer-based actuators (with/without internal chambers and reinforcements) have been developed for achieving programmed displacements such as contraction, bending and twisting [178, 200, 218, 225, 226] (Fig. 7g). Connolly et al. provided a solution to derive reinforcement configurations based on desired tip trajectories [256]. A different design approach by Pylatiuk et al. proposed a wearable system using an actuator inspired by spider legs [228, 229]. Some of these systems have been developed for at-home rehabilitation with a portable actuation pack for hydraulic fluid/air canisters and controllers [210, 226] (Fig. 7h). However, compared to cable-driven systems, pneumatic and hydraulic systems suffer from limitations such as lower control bandwidth, slow response time, and restricted portability due to their tethering to air compressors or tanks (though smaller portable canisters are an option) [223, 226]. Soft pneumatic actuators also face challenges with low output forces, low functional bandwidth in the case of PAMs (though force output is significantly large), noise, and safety concerns (from sudden pressure release due to leaks/bursts) [183]. PAMs, like most other novel soft actuation methods, exhibit hysteresis and significant nonlinear behaviour, necessitating the development of appropriate control strategies particularly to achieve accurate joint trajectory tracking. Relatively straightforward as well as more sophisticated modelling methods including echo state networks [230] have been used to approximate the system dynamics for better behaviour prediction. Recently, hybrid models such as combined pneumatic and electric actuators have been used to bring together advantages of both actuation principles such as accuracy, force amplitude and backdrivability. Actuation systems must also manage extreme temperature variations, vibration isolation, radiation-hardening of controller electronics and other space-specific limitations.

Artificial actuators can be avoided or complemented by eliciting muscle contractions with FES/NMES. By using natural muscle power elicited by electrical stimulation, it is possible to reduce the size and weight of the overall system that can be worn under clothing [132, 187, 228, 231]. FES and stimulation, in general, are particularly important for space physiology since they can be used to counteract muscle atrophy. Because of miniaturization of the electronics, FES systems can be fully wearable. Nonetheless, FES still suffers from many limitations such as poor controllability, high levels of muscle fatigue, discomfort, and erratic output [180, 181, 232]. Poor controllability stems from the neuromusculoskeletal system being a highly non-linear and time-variant system, and hence, the same input stimulation can result in drastically differing behaviours at different instances, and sustained stimulation results in fatigue and discomfort for the user. Hence, sophisticated modelling and control of the combined human-suit system becomes necessary. Furthermore, FES could be coupled with actuation technologies such as motors, to facilitate fine control. The same principle could also be leveraged during rehabilitation and physical therapy [233]. The technology has the potential to serve both as an assistive system and as a means to maintain muscle tone and prevent atrophy.

With research in soft robotics gaining attention and traction, researchers have developed several soft actuation technologies to deploy in wearable robotics. Interdisciplinary research in basic sciences, materials and other fields have led to novel actuation technologies such as shape memory materials [234, 235] ionic/electronic electro-active polymers (EAPs) [236], dielectric elastomers [237], twisted nylon coil artificial muscles (twisted string actuators) [238], dielectrophoretic liquid zipping (DLZ) actuators [239], hydraulically amplified self-healing electrostatic (HASEL) actuators [240], magnetorheological (MR) and electrorheological (ER) fluids [185, 241]. The above and other concepts whether it be fluidic fabric muscle sheets, liquid crystal elastomers, magnetoactive soft materials, thermally responsive hydrogels etc. while at low TRL for real-world translation in wearable robotics, could soon find incorporation for rehabilitation, assistance and stability/support applications for both space and terrestrial applications. However, before this can be achieved, challenges and limitations need to be overcome such as precisely modelling and controlling for the nonlinear behaviour of most of these actuators amongst others. Also, concept-specific limitations range from the slow response of shape memory materials and twisted string actuators to oil retention and encapsulation of DLZ actuators, or low force capabilities of electroactive polymers and dielectric elastomers. Advantages and disadvantages of both traditional and novel actuation mechanisms are tabulated below.

TABLE I: ACTUATION PRINCIPLES IN WEARABLE SYSTEMS

| Actuation Principle | Brief Description | Pros | Cons |
|---|---|---|---|
| **Passive Elements (springs, dampers, gyros, …)** | Forces provided proportional to quantities such as displacement, velocity, etc. | Low complexity, predictable behaviour, no power source needed | Lack of ability to change force profiles on-demand |
| **Motors (DC/BLDC)** | Electromechanical actuators for direct-drive or transmitted using cable-driven or other systems | Reliable, controllable, compact, widely available | High stiffness, potential safety issues during failure, higher weight compared to pneumatics |
| **Pneumatic Actuators** | Use compressed air for force generation, includes pneumatic artificial muscles (PAMs) | Lightweight, high power-to-weight ratio | Requires air compressors/tanks, slow response time, noise, safety concerns from leaks, nonlinear characteristics, hysteresis in case of PAMs |
| **Hydraulic Actuators** | Fluid-driven actuators capable of providing high force | High force generation, efficient, capable of precise control | Heavy, noisy, prone to fluid leakage, complex maintenance |
| **Series Elastic Actuators (SEAs)** | Motors coupled with elastic elements for compliance | High compliance, safer interaction, better shock absorption | Narrow functional bandwidth, limited use in high-speed tasks |
| **Functional Electrical Stimulation (FES)** | Elicits muscle contractions through electrical stimulation | Compact, leverages natural muscle power, can be energy-efficient | High fatigue, discomfort, poor controllability, requires sophisticated control |
| **Shape Memory Alloys (SMA)** | Materials that change shape upon heating or electrical activation | Compact, lightweight, simple design | Slow response time, low force generation, limited functional lifespan |

| Electroactive Polymers (EAPs) | Polymers that change shape in response to electrical stimulation | Flexible, lightweight, energy-efficient, silent | Low force generation, limited durability, complex control, nonlinearities, high voltages required |
|---|---|---|---|
| Dielectric Elastomers | Elastomers that deform under electrical activation | Fast response, programmable, high efficiency | Low force generation, requires pre-stretching, requires high voltage, nonlinearities |
| Twisted String Actuators | Coiled nylon or other materials twisted to produce contraction | Lightweight, compact, high force-to-weight ratio | Slow response, hysteresis, complex control mechanisms |
| HASEL Actuators | Electrohydraulic transducers with muscle-like motion | Muscle-mimetic, programmable, versatile applications | Oil retention issues, encapsulation challenges, sensitive to temperature variations |
| DLZ Actuators | Dielectrophoretic actuators | High control bandwidth and power-to-weight ratio | Requires high voltage, oil retention and encapsulation, nonlinearities |
| Magnetorheological/ Electrorheological Fluids | Fluids whose viscosity changes under magnetic or electric fields | High adaptability, can be used for variable stiffness | Heavy, requires active control systems, limited portability, can only dissipate energy |
| Fluidic Fabric Muscle Sheets | Flexible sheets integrated with fluid-driven actuation | Lightweight, conforms to shapes, energy-efficient | Low force generation, limited robustness |

## 5.3 Wearable Sensing Technologies

Sensors that monitor psycho-physiological state, musculoskeletal adaptations, human-spacesuit interactions, and countermeasure effectiveness are critical to astronaut well-being and operational success, alongside facilitating human-robot synergy in wearable robots [2, 181, 243]. The strict operational demands by space agencies such as NASA, ESA, JAXA, ROSCOSMOS, etc., necessitate compact, user-friendly, rugged devices with long battery life and FDA-approved clinical support [244]. Additionally, these sensors must resist the radiation beyond Earth's atmosphere, requiring active electronics developed using radiation hardening techniques and shielding—particularly within the South Atlantic Anomaly (SAA). This subsection aims to briefly discuss different sensing modalities loosely along the lines of: i) physiological effects of adaptations and injuries, and ii) biomechanical sensing and detecting user intent.

### 5.3.1 Injury, Fatigue, Adaptation & Physiological Monitoring

While the *Extravehicular Mobility Unit* (EMU) suit and glove provide life support in outer space's harsh environment, their prolonged use often results in fatigue and injuries, particularly to fingertips and fingernails. The challenging environment and reduced efficacy of medications in space necessitate preventative monitoring. Sensing modalities employed for this purpose range from laser Doppler flowmetry probes to piezoresistive sensor strip arrays, humidity sensors, and thermocouples [95, 245]. Additionally, multi-sensory glove-based approaches involving galvanic skin response and barometric pressure sensors have been used to assess skin moisture/perspiration and transient pressure changes during dynamic tasks [246]. Other approaches include the electromagnetic skin patch with radio frequency resonant spiral proximity sensor [93] proposed for assessing the distance between the suit, the *Liquid Cooling and Ventilation Garment* (LCVG) and the skin. However, these sensors are affected by their proximity to the metal in the HUT (Fig.8d).

Other sensing technologies for monitoring body-suit-environment interaction can measure forces starting from tactile/haptic ranges to get a measure of discomfort and potential for injury. Composite material-based strain/pressure sensors embedded in suits are a promising sensing method, owing to their versatility and fabrication using techniques such as extrusion-based 3D printing, laser carbonisation, injection moulding, and/or stencils printing [249, 253–257] (Fig. 8a-c). These sensors typically employ a polymer matrix and conductive filler, facilitating the sensing of suit-body-environment interactions and using the data to optimise human-suit(robot) ergonomics and controller design [243]. In [90], human-suit interaction was assessed using a pressure sensing mat on the shoulder and custom pressure sensors along the arm.

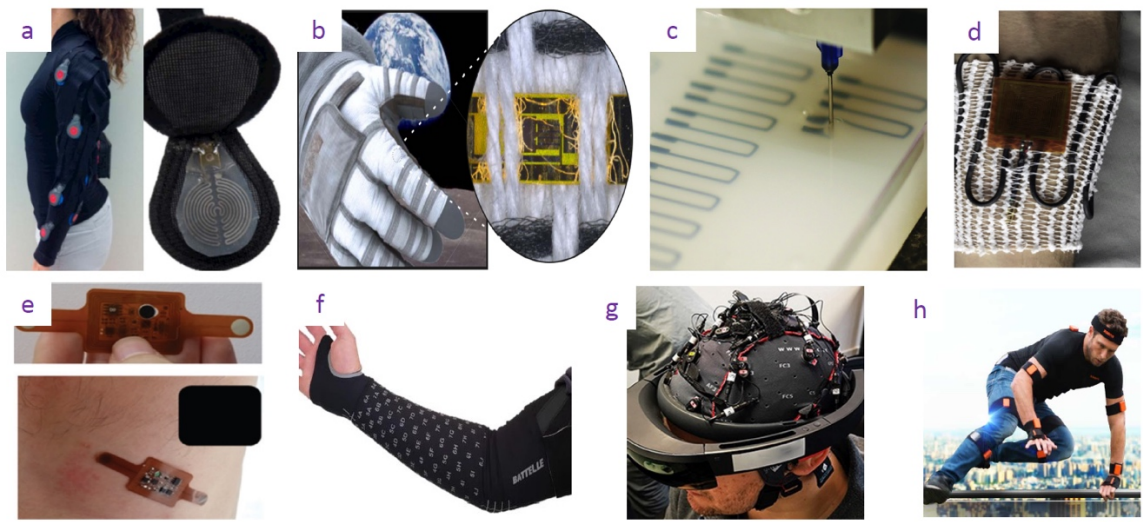

**Figure 8:** An overview of wearable sensing systems developed for terrestrial and space applications. **Top Row: a).** A textile-integrated liquid metal-based resistive pressure sensor array for spacesuit dynamics [247]; **b).** Textile-integrated InGaZnO (IGZO) thin-film transistors for space applications [248]; **c).** Extrusion 3D printing of directly embedded resistive strain sensors [249]; **d).** Electromagnetic resonant spiral proximity sensors for monitoring shoulder joint clearance in space suit for injury prevention [93]; **Bottom Row**: **e).** NFC-powered flexible chest patch for fast assessment of cardiac, hemodynamic, and endocrine parameters [250]; **f).** a wearable high density EMG sleeve developed by Battelle Memorial Institute [251]; **g).** subject wearing AR headset (Microsoft HoloLens) and an EEG headset [252]; **h).** the XSens IMU suit developed by Movella ©Movella.

These versatile soft sensors utilise microfluidic channels in an elastomer filled with liquid metal (Gallium-Indium-Tin eutectic) (Fig.7a). Radial/circumferential, shear and normal strains could be measured, making this approach suitable for both generalised and application-specific interaction/injury monitoring and to assess body kinematics or dynamics [90, 247] (Fig.8a). Thin film technology based on clean room-based microfabrication techniques, has found applications in this field (Fig. 8b) due to the potential realisation of high performance and highly miniaturised flexible and stretchable devices and the ability to co-integrate readout electronics with sensors such as strain and electrophysiological sensors [248]. Among the various technologies explored, indium gallium zinc oxide (IGZO) based transistors hold great promise for the realization of flexible transistors [248]. Recently, e-skin sensing equipment has been developed by Song et al. [258], which can measure temperature and pressure simultaneously, allowing the monitoring of injury or stress. However, it should be noted that these systems can have lower reliability/robustness and require significant improvement in TRL before it can be adopted for space applications.

In addition to being informed about the onset and extent of injuries, an understanding of general physical and mental well-being and the impact of different maladaptations in space is critical. Wearable multiparametric sensing systems were developed for this purpose, such as the Canadian Space Agency's Astroskin Bio-Monitor system [259] (measuring activity level, breathing rate, blood oxygen saturation, skin temperature, ECG, and systolic BP) and a *polysomnography* (PSG) system for monitoring sleep quality [260] (measuring EMG, EEG, ECG, *electrooculography* (EOG), thoracic movements, and airflow).

Similarly, getting a quantitative understanding of the extent of musculoskeletal and other adaptations in (near) real-time through the measurement of electrolytes, proteins, and other biomarkers based on research reviewed in the previous sections could be vital for individual-specific tuning of nutrition, supplementation, and exercise regimen. While not developed specifically for space applications, several multiparametric sensing devices for comprehensive analysis of body fluids, such as sweat, including amperometric (monitoring metabolites like glucose), impedimetric (monitoring biomarkers such as stress hormones like cortisol), potentiometric (measuring electrolytes such as pH, calcium, etc.) and/or bioimpedance (monitoring tissue hydration, galvanic skin response, and tissue ischemia), could be further optimised and tailored for obtaining real-time feedback on musculoskeletal adaptations (Fig. 8e) [250, 261–264]. Many of these systems are still lab-based and need to see real-world product development focussed more on improving robustness and reliability.

Beyond health and injury, monitoring physical and mental fatigue is also critical, especially from the perspective of mission-critical and dangerous IVA/EVA activities. Muscle fatigue, evaluated using sEMG and Mosso's ergograph (Fig. 8f), could be particularly useful, especially during labour-intensive IVA/EVA activities [218]. Technologies integrating multiple sensing modalities, such as brain-computer interfaces (BCIs) integrated with augmented/virtual reality (AR/VR) (Fig. 8g), computer vision and eye-tracking, can not only provide context awareness but also real-time monitoring and adaptive instructions, drastically improving quality and efficiency during critical IVA/EVA tasks [265–268].

### 5.3.2 Sensing Kinematics, Dynamics & Detecting Intent

Having wearable robotic systems worn by astronauts purely as an exercise countermeasure will not leverage all the technology's capabilities. Therefore, if astronauts intend to use wearable robotic systems for active assistance during IVA/EVA tasks, misinterpretation/ delays in understanding user intent would result in unnatural compensations, increased mistakes, frustration, and eventual disuse of the technology. Therefore, apart from the monitoring applications discussed previously, sensing body-suit-environment state, interaction, kinematics, dynamics, and intent in real-time becomes vital [150, 242, 243].

Control inputs in wearable robotics have evolved from simple analogue [205], and expiration switches [202, 271] to advanced electrophysiological measurements based on EMG (Fig. 8f) and *brain-computer interfaces* (BCIs) (Fig. 8g). Electrophysiological measurements can be superior as they can even detect the onset of movement. Surface EMG is valued for its non-invasiveness, ease-of-use and compatibility with wearables while offering insight into neural intent [269, 272]. Kuroda et al. [242] have successfully utilized myoelectrical signals generated by muscle contractions for both sensing and actuating hand control for tactile interaction. This and similar other systems enable the intuitive control of devices by translating natural muscle activity into mechanical movements. Advances have incorporated biomechanical and neuromusculoskeletal models for smoother control [273, 274], high-density EMG (HD-EMG) data acquisition (Fig. 8f) [251], and novel signal processing techniques [269, 275–278] for intuitive control of orthotics and prosthetics. Wearable systems for monitoring muscle activation in astronauts have been developed and tested [279, 280].

BCIs are gaining attention in wearable robotic control, recording EEG or metabolic fNIRS changes [281]. They have been used to control robots through motor imagery [187], visually-evoked potentials [282], and P300 signals [283] protocols; however, they have not demonstrated accurate real-time control yet. BCIs, combined with other sensing modalities like eye-tracking and computer vision, could aid in context awareness and more accurate user intent detection and cognitive workload monitoring [267] (Fig. 8g). While real-time BCI control remains a challenge, promising research efforts based on issuing higher-order commands than low-/medium-level control [284], employing fast-switch-based methods [285], and utilising sensory predictions generated by forward models [286] are being used to inform robot movements and move closer to achieving real-time control.

In tandem with muscle and neural activity signals, various strain, force, and kinematic sensing approaches could offer vital insight into the human-robot system state, interactions, and user intent and regulate controller input [186, 211, 234, 287, 288]. Joint angle [226, 289], velocity [290], and acceleration measurements are frequently employed, occasionally in conjunction with joint torque metrics [291, 292]. The ubiquitous usage of IMUs (Fig. 8h) in numerous real-world systems, including space-based wearables, underscores their effectiveness [143, 144, 214, 216, 293]. Textile-based wearable sensors offer an innovative alternative for kinematics and physiological monitoring. Combined with statistical models and neural networks, they enable accurate prediction of complex body movements, such as torso/lumbar shape/posture [92] and multi-DoF movements [294].

For intuitive human-robot interaction, force-sensing technologies are essential to get a sense of both dynamics and human-robot-environment interaction. They provide valuable metrics to controllers, with tactile [202, 220], capacitive [223, 256], and resistive sensors [197, 257] tracking user intent and generating appropriate movements and forces. Inductive [295] and deformation-based sensors based on load cells [296] monitor interaction forces, forming key input for the wearable robot's closed-loop controller. Conductive composite, liquid metal, and other strain/pressure sensing modalities [247, 249, 253–257] (Fig. 8a,c) discussed in the previous sub-section could also be used for the sensing interaction and obtaining tactile/haptic feedback. Context awareness using data from cameras and eye trackers, potentially integrated within AR/VR headsets (Fig.8g), used alongside neural/muscle activity signals could help identify user intent more accurately [252, 297]. Different sensing modalities and their advantages/disadvantages are tabulated below.

TABLE II: SENSING MODALITIES IN WEARABLE SYSTEMS

| Sensing Modality | Brief Description | Pros | Cons |
|---|---|---|---|
| Laser Doppler Flowmetry Probes | Measures blood flow velocity using laser light reflection. | Non-invasive, high sensitivity. | Requires precise positioning. Wearability, miniaturization and power consumption can be issues. |
| Humidity Sensors | Monitors skin moisture levels for perspiration assessment. Typically capacitive or resistive. | Simple, effective for moisture detection. | Limited accuracy under dynamic conditions due to response time limitations. |
| Thermocouples | Detects temperature variations to monitor physiological responses. | High accuracy, widely used. | Slower response time in varying conditions. |
| Galvanic Skin Response Sensors | Measures skin conductance to monitor perspiration and stress levels. | Tracks psycho-physiological states. | Unreliable due to sensitivity to external environmental factors. |
| Barometric Pressure Sensors | Detects transient pressure changes during dynamic tasks. | Compact, easy to integrate. | Limited to localized pressure sensing. |
| Electromagnetic Proximity Sensors | Tracks skin-to-suit distance using RF resonance. | Provides spatial clearance information. | Affected by metal proximity interference. |
| Thin-Film IGZO Transistors | Flexible, stretchable thin film device for strain and electro-physiological sensing. | Co-integration with electronics, high miniaturization, bulk fabrication, high performance. | Limited durability under high strain, complex and expensive fabrication. |
| Piezoresistive Sensor Strip Arrays | Measures pressure changes based on resistivity variation in strips. | Low-cost, robust, reliable. | Limited resolution for small-scale forces. Hysteresis is an issue. |
| Composite Strain/Pressure Sensors | Embedded in suits, measures interaction forces using polymer matrix and conductive fillers. | Versatile, customised and affordable fabrication possible. | Complex to fabricate, hysteresis, and transiency in sensor baseline. |
| Microfluidic Liquid Metal Sensors | Measures strain with liquid metal-filled elastomer channels for normal and shear forces. | High flexibility, multi-directional. | Complex fabrication and integration into textiles. |
| Textile-Based Strain Sensors | Fabric-integrated sensors to measure posture and complex movements. | Flexible, wearable under clothing. | Limited sensitivity for fine movements. |
| Tactile and Resistive Sensors | Measures interaction forces for haptic and tactile feedback in wearable robots. | Cost-effective, reliable. | Limited precision in dynamic settings. |
| Capacitive Force Sensors | Measures user interaction forces through capacitance variations. | High accuracy, low hysteresis. | Sensitive to environmental changes. |
| E-Skin Sensors | Simultaneously measures temperature and pressure for injury/stress monitoring. | Multi-sensing capability, compact. | Requires specialized materials. |
| Astroskin Bio-Monitor System | Monitors multiple parameters: activity, breathing rate, blood oxygen, ECG, BP, and skin temp. | Multi-parametric, compact system. | Radiation shielding needed for space. |
| Polysomnography (PSG) System | Monitors sleep quality using EMG, EEG, ECG, EOG, thoracic movements, and airflow. | Comprehensive physiological monitoring. | Bulky, less portable. |
| Surface EMG (sEMG) | Detects muscle activation signals to measure fatigue and intent. | Non-invasive, high usability. | Motion artifacts and sensitive to changing surface conductivity, requiring calibration. |
| Brain-Computer Interfaces (BCIs) | Records EEG or metabolic fNIRS changes for monitoring neural intent and workload. | Enables intent detection, hands-free. | Limited real-time accuracy, noisy signals. |

| Inertial Measurement Units (IMUs) | Tracks body kinematics (angle, velocity, acceleration) in real-time for motion monitoring. | Compact, widely adopted. | Signal drift over time. |
|---|---|---|---|
| Eye-Tracking Sensors (AR/VR Integrated) | Tracks gaze direction for context awareness and intent detection in AR/VR applications. | Enables real-time interaction feedback. | Requires integration with headgear. |

# 6. Discussion, Perspectives for Future Research and Development

This review is intended for researchers and designers aiming to develop the next generation of wearable countermeasures to mitigate the harsh stressors of outer space. We have examined the physiological adaptations of the human body to the space environment, discussed existing countermeasures, and explored the potential of wearable technologies to minimise harmful adaptations.

Current research in space physiology, particularly musculoskeletal adaptations, provides valuable insights into bone density loss, muscle atrophy, and systemic physiological changes caused by prolonged microgravity exposure. Studies conducted pre- and post-flight, especially those leveraging sophisticated imaging modalities and biochemical assays on the ISS, have laid a strong foundation for understanding these complexities. While countermeasures such as resistive exercise, nutritional strategies, and pharmacological interventions have shown promise in mitigating these adaptations, they remain only partially effective, often bulky, time-intensive, and resource-demanding. These limitations pose significant challenges for future missions aboard compact spacecraft. Exercise is the cornerstone of current countermeasures, with systems like CEVIS, TVIS, iRED, and ARED installed on the ISS and earlier missions such as Mir and Skylab. Novel solutions, including artificial gravity, are also under evaluation. However, the constraints of weight, size, and payload capacity demand sophisticated, lightweight, modular, and compact alternatives.

Wearable exercise countermeasures hold significant promise in overcoming these limitations. Concepts such as the Pingvin suit and its modern successors, including the GLCS and V2, have demonstrated improved performance. Space agencies like NASA, ESA, JAXA and ROSCOSMOS have also explored wearable robotic systems, such as exoskeletons, for rehabilitation and assistance. Future astronaut suits could evolve into intelligent "second skins," integrating wearable actuation, sensing, computing, and intelligence. Advances in materials science, battery technology, and embedded systems could enable these suits to provide real-time monitoring, energy harvesting, and intuitive physical and cognitive assistance.

Materials Science
Materials science plays a pivotal role in enabling the next generation of wearable technologies and should be a key focus area. It provides the platform on which other technologies whether it be in human-machine interfacing, controls or AI can build on. Innovations in materials are essential for advancements in battery technology, spacesuit design, sensors, actuators, and electronics. Spacesuit embodiments must address thermoregulation, pressure management (mechanical counterpressure versus pressurized suits), comfort, dexterity, and ergonomics. Research on power sources and battery technologies is equally critical to developing practical, unencumbered, and efficient wearable systems. Space environments present unique challenges such as extreme temperature variations, vacuum, and radiation. Hydrogen-rich polymers, boron nitride nanotubes, and graphene and other promising materials for radiation shielding and mechanical resilience should be a focus area for researchers. Also, research into radiation-hardening of electronics using wide bandgap semiconductors (like silicon carbide and gallium nitride) and other methods is a crucial focus are for ensuring reliable operation in extreme conditions.

Sensing and Actuation Technologies
Novel sensing modalities, including electrochemical, myoelectric, kinematic, and environmental sensors, are critical for real-time monitoring and decision-making. High-quality data from these sensors can facilitate better operational decisions, human-robot interactions, and adaptive control of robotic platforms. Soft robotics and soft sensing technologies offer exciting avenues for innovation, though challenges such as modelling anc controlling nonlinear behaviour, high-voltage requirements, material durability, slow response times and others must be addressed to transition from laboratory research to high-TRL (technology readiness level) systems. In space applications, sensors for biomarker analysis, bone loss monitoring, and injury detection are priorities. Similarly, the development of actuators providing reliable assistance, resistance for physical therapy, and feedback for tasks such as teleoperation should also be a key focus. Neuromusculoskeletal modelling and adaptive control algorithms will be crucial for achieving seamless integration of wearable robotic platforms with astronauts' natural movements.

AI & Digital Twins
The exponential growth of AI presents opportunities to model materials, mechanisms, and biological systems. Astronauts' extensive pre-, during, and post-flight testing has generated a wealth of data that could be used to create highly accurate digital twins, both neuromusculoskeletal and other systems. These twins could represent various physiological systems and adapt in real time using data from onboard sensors. AI-powered systems could analyse this data and be a decision support system for predictive and reactive medical/diagnostic interventions, enable personalised exercise regimen and adaptive controllers in robotic systems resulting in improved human-machine interaction, help finely tune nutrition and pharmacological protocols, and lead to better cognitive and physical health. This makes this research area an extremely vital area for both development and testing of new systems as well as their robust deployment on space missions.

Human-Robot Interaction
This field of research is pivotal for developed wearable robotic technologies to become a natural and intuitive extension of the wearer. Insights and models derived from obtained data hold unparalleled potential for enabling seamless, intuitive, and natural human-robot interaction, both cognitive and physical. For example, the development of robust and accurate digital twins simulating neurophysiology, neuromechanics, and other systems represent a paradigm shift in the development and control of wearable robotic and other platforms. Challenging actuation methods, such as FES/NMES, could become far more predictable and controllable through these advancements. Such progress would enable shared control strategies that harness the strengths of FES/NMES alongside traditional and novel actuation approaches, resulting in highly optimized and efficient wearable robotic platforms.

Innovative human-machine interaction strategies, leveraging technologies such as HD-EMG, ultrasound, IMUs, EEG, and other sensing modalities—individually or in fusion—could further facilitate intuitive and natural control. However, substantial research is needed, including AI-based algorithm development, to overcome practical challenges such as the frequent need for calibration and complex amplification electronics.

Beyond assistance or therapy, modelling the interactions between humans, robotic suits, and the environment can pave the way for providing haptic, tactile, and other forms of biofeedback. This is particularly important for astronauts, as the loss of haptic feedback and dexterity, especially when wearing pressurized gloves during EVA is well-documented. Research into mechatronic systems capable of delivering such feedback is vital to create a more seamless and natural user experience.

Additionally, exploring methodologies for shared human-human and human-robot control in platforms used for regular maintenance and IVA/EVA is critical. These advancements could significantly reduce physical strain, mitigate cognitive overload, and enhance overall task performance. From the cognitive perspective, employing multiple sensing modalities, such as various BCI systems, can provide a highly accurate assessment of mental acuity, fatigue, and cognitive load, helping to prevent accidents.

Realizing a wearable robotic system that integrates all these advancements could lead to a future where astronauts achieve superior physical and cognitive performance while wearing minimalistic wearable robotic systems that would feel as natural as clothing. Such systems would not only optimize their physical health and mental well-being during missions but also provide lasting benefits long after the mission's completion.

<u>Challenges and Limitations</u>
Despite advances, significant challenges remain. Most space physiology research has been conducted in low Earth orbit limiting the applicability of learnings to extended missions or deep-space environments. Additionally, limited sample sizes, disparate methodologies, gender imbalances and an overall lack of generalisability amongst findings from different space agencies needs careful consideration. Establishing standardised research frameworks and addressing these biases will be essential for advancing countermeasure development. Furthermore, as opposed to academic research, research for space is not widely disseminated making it challenging for private companies and academic researchers to gain access to the latest technologies and to contribute more effectively, and should be a focus area from a policy perspective.

<u>Future Applications and Broader Impact</u>
Space R&D has historically driven innovations with far-reaching terrestrial applications, from medical devices to everyday technologies. Wearable countermeasures developed for astronauts could translate into systems for at-home rehabilitation, personal fitness, and assistance for the elderly. By addressing the challenges faced by astronauts, researchers can also develop solutions that enhance physical and cognitive abilities, benefiting diverse populations on Earth.

Through this review, we aim to provide a comprehensive foundation for advancing wearable countermeasures, emphasising both the physiological challenges and technological opportunities for future space and terrestrial applications.

**Acknowledgements**

We would like to thank Dr. Fani Deligianni, Dr. Benny Lo, and Prof. Guang-Zhong Yang for their early discussions and comments on this topic. This work was supported in part by the U.K. Engineering and Physical Science Research Council (EPSRC) with the Future AI and Robotics Hub for Space (FAIR-SPACE, EP/R026092/1) and the Non-Invasive Single Neuron Electrical Monitoring (NISNEM, EP/T020970/1) projects, as well as the European Union with the Horizon 2020 ICT CONnected through roBOTS (CONBOTS) project (871803).

**Conflict of Interest:** The author(s) declare(s) that there is no conflict of interest regarding the publication of this article.

**Data Availability:** Submission of a manuscript to *Cyborg and Bionic Systems* implies that the data is freely available upon request or has been deposited to an open database, like NCBI. If data are in an archive, include the accession number or a placeholder. Also, include any materials that must be obtained through an MTA.

111. Rittweger J, Beller G, Armbrecht G, et al. Prevention of bone loss during 56 days of strict bed rest by side-alternating resistive vibration exercise. Bone 2010;46:137–47.

112. Hurst C, Scott JPR, Weston KL, and Weston M. High-Intensity Interval Training: A Potential Exercise Countermeasure During Human Spaceflight. Frontiers in Physiology 2019;10.

113. Jones TW, Petersen N, and Howatson G. Optimization of Exercise Countermeasures for Human Space Flight: Operational Considerations for Concurrent Strength and Aerobic Training. Frontiers in Physiology 2019;10.

114. Caplan N, Weber T, Gibbon K, Winnard A, Scott J, and Debuse D. The Functional Re-adaptive Exercise Device: preferential recruitment of local lumbopelvic muscles.

115. Kozlovskaya IB and Grigoriev AI. Russian system of countermeasures on board of the International Space Station (ISS): the first results. Acta Astronautica 2004;55:233–7.

116. Bellisle R and Newman D. Countermeasure suits for spaceflight. In: 2020 International Conference on Environmental Systems. 2020.

117. Newman D, Hoffman J, Bethke K, Blaya J, Carr C, and Pitts B. Astronaut bio-suit system for exploitation class missions. NIAC Phase II Final Report 2005.

118. Duda KR, Vasquez RA, Middleton AJ, et al. The variable vector countermeasure suit (v2suit) for space habitation and exploration. Frontiers in Systems Neuroscience 2015;9:55.

119. NASA: Artemis. NASA. URL: https://www.nasa.gov/specials/artemis/index.html (visited on 04/16/2023).

120. China space station: What is the Tiangong? 2022.

121. Axiom Space — World's First Commercial Space Station. URL: https://www.axiomspace.com/ (visited on 04/16/2023).

122. Bigelow Aerospace - B330. Bigelow Aerospace. URL: http://www.bigelowaerospace.com/b330 (visited on 04/16/2023).

123. Scott JPR, Weber T, and Green DA. Introduction to the Frontiers Research Topic: Optimization of Exercise Countermeasures for Human Space Flight – Lessons From Terrestrial Physiology and Operational Considerations. Frontiers in Physiology 2019;10.

124. Mayr W, Bijak M, Girsch W, et al. MYOSTIM-FES to Prevent Muscle Atrophy in Microgravity and Bed Rest: Preliminary Report. Artificial Organs 1999;23:428–31.

125. Duvoisin MR, Convertino VA, Buchanan P, Gollnick PD, and Dudley GA. Characteristics and preliminary observations of the influence of electrostimulation on the size and function of human skeletal muscle during 30 days of simulated microgravity. Aviation, Space, and Environmental Medicine 1989;60:671–8.

126. Shiba N, Matsuse H, Takano Y, et al. Electrically Stimulated Antagonist Muscle Contraction Increased Muscle Mass and Bone Mineral Density of One Astronaut - Initial Verification on the International Space Station. PLOS ONE 2015;10. Publisher: Public Library of Science:e0134736.

127. Reidy PT, McKenzie AI, Brunker P, et al. Neuromuscular Electrical Stimulation Combined with Protein Ingestion Preserves Thigh Muscle Mass But Not Muscle Function in Healthy Older Adults During 5 Days of Bed Rest. Rejuvenation Research 2017;20. Publisher: Mary Ann Liebert, Inc., publish- ers:449–61.

128. Dirks ML, Wall BT, Snijders T, Ottenbros CLP, Verdijk LB, and Loon LJC van. Neuromuscular electrical stimulation prevents muscle disuse atrophy during leg immobilization in humans. Acta Physio- logica 2014;210. _eprint: https://onlinelibrary.wiley.com/doi/pdf/10.1111/apha.12200:628–41.

129. NASA. Muscle Stimulation Technology. Spinoff 1997 1997. NTRS Document ID: 20020076193 NTRS Research Center: Goddard Space Flight Center (GSFC).

130. NASA Life Sciences Portal: Record Viewer.

131. Ives JC. Multi-User Facilities for Human Physiology. 1996;385. Conference Name: Space Station Utilisation ADS Bibcode: 1996ESASP.385...77I:77.

132. Maffiuletti NA, Green DA, Vaz MA, and Dirks ML. Neuromuscular Electrical Stimulation as a Poten- tial Countermeasure for Skeletal Muscle Atrophy and Weakness During Human Spaceflight. Frontiers in Physiology 2019;10:1031.

133. Hargens AR, Bhattacharya R, and Schneider SM. Space physiology VI: exercise, artificial gravity, and countermeasure development for prolonged space flight. European Journal of Applied Physiology 2013;113:2183–92.

134. Lackner JR and DiZio P. Artificial gravity as a countermeasure in long-duration space flight. Journal of Neuroscience Research 2000;62:169–76.

135. Clément GR, Charles JB, and Paloski WH. Revisiting the needs for artificial gravity during deep space missions. Reach 2016;1:1–10.

136. Kotovskaya AR. The problem of artificial gravity: The current state and prospects. Human Physiology 2010;36:780–7.

137. Kaderka J, Young LR, and Paloski WH. A critical benefit analysis of artificial gravity as a microgravity countermeasure. Acta Astronautica 2010;67:1090–102.

138. Young LR, Hecht H, Lyne LE, Sienko KH, Cheung CC, and Kavelaars J. Artificial gravity: Head movements during short-radius centrifugation. Acta Astronautica 2001;49:215–26.

139. Adeli suit - Wikipedia.

140. Waldie JM and Newman DJ. A gravity-loading countermeasure skinsuit. Acta Astronautica 2011;68:722–30.

141. Attias J, Philip ATC, Waldie J, Russomano T, Simon NE, and David AG. The Gravity-Loading coun- termeasure Skinsuit (GLCS) and its effect upon aerobic exercise performance. Acta Astronautica 2017;132:111–6.

142. Carvil PA, Attias J, Evetts SN, Waldie JM, and Green DA. The Effect of the Gravity Loading Countermeasure Skinsuit Upon Movement and Strength. Journal of strength and conditioning research 2017;31:154–61.
Manuscript Template Page 39 of 50